\documentclass{article}
\pdfoutput=1

\usepackage[nonatbib, final]{neurips_glfrontiers_2023}

\usepackage{tikz}
\usepackage{subcaption}
\usetikzlibrary{shapes,arrows,positioning}
\usetikzlibrary{fit}
\usepackage{amsmath}
\usepackage{enumitem}
\usepackage[
    bibstyle=ieee,
    citestyle=numeric,
    isbn=true,
    doi=true,
    sorting=none,
    url=true,
    bibencoding=utf8,
    backend=bibtex,
    natbib=true
]{biblatex}
\usepackage{amsfonts}
\usepackage{booktabs}
\usepackage{colortbl}
\usepackage{tabularray}
\newcommand\blfootnote[1]{\begingroup
  \renewcommand\thefootnote{\arabic{footnote}}\addtocounter{footnote}{-1}\endgroup
}

\usepackage[utf8]{inputenc} \usepackage[T1]{fontenc}    \usepackage{hyperref}       \usepackage{url}            \usepackage{booktabs}       \usepackage{amsfonts}       \usepackage{nicefrac}       \usepackage{microtype}      \usepackage{xcolor}

\title{Learning Genomic Sequence Representations using Graph Neural Networks over De Bruijn Graphs}

\author{Kacper Kapu\'sniak\thanks{The author is now affiliated with the University of Oxford: \href{mailto:kacper.kapusniak@keble.ox.ac.uk}{kacper.kapusniak@keble.ox.ac.uk}.} \\
    ETH Z\"urich, Switzerland\\
    \texttt{kkapusniak@student.ethz.ch} \\
  \And
    Manuel Burger \\
    ETH Z\"urich, Switzerland\\
    \texttt{manuel.burger@inf.ethz.ch} \\
  \And
    Gunnar R\"atsch \\
    ETH Z\"urich, Switzerland\\
    \texttt{raetsch@inf.ethz.ch} \\
  \And
    Amir Joudaki \\
    ETH Z\"urich, Switzerland\\
    \texttt{amir.joudaki@inf.ethz.ch} \\
}

\begin{document}

\maketitle

\begin{abstract}
  The rapid expansion of genomic sequence data calls for new methods to achieve robust sequence representations. Existing techniques often neglect intricate structural details, emphasizing mainly contextual information. To address this, we developed k-mer embeddings that merge contextual and structural string information by enhancing De Bruijn graphs with structural similarity connections. Subsequently, we crafted a self-supervised method based on Contrastive Learning that employs a heterogeneous Graph Convolutional Network encoder and constructs positive pairs based on node similarities. Our embeddings consistently outperform prior techniques for Edit Distance Approximation and Closest String Retrieval tasks. \footnote[1]{Code Available at \href{https://github.com/ratschlab/genomic-gnn.git}{https://github.com/ratschlab/genomic-gnn.git}} \end{abstract}

\section{Introduction}

Genomic sequence data is expanding at an unparalleled pace, necessitating the development of innovative methods capable of providing accurate and scalable sequence representations~\cite{genomics_more_data}. These representations are foundational for many computational biology tasks, ranging from gene prediction to multiple sequence alignment~\cite{genomic_ml_applications_review}.
The computational biology community has adopted methods from Natural Language Processing (NLP), such as Word2Vec and Transformers, to improve the representation of genomic sequences~\cite{dna2vec,dnabert,kmer2vec,metagenome2vec,word2vec_genomics}. These NLP-based approaches are adept at capturing context within the sequence, a vital aspect where the semantics of words often outweigh the precise letters composing them.

To capture structural nuances, one might consider character-level n-gram models. However, a uniform representation of each n-gram across all sequences can oversimplify the problem, and on the other hand, applying techniques like transformer-based models on n-grams can escalate computational demands. Consequently, these methods may overlook nuanced k-mer variations essential for comprehending single-nucleotide polymorphisms (SNPs) and other minor sequence changes. These SNPs influence disease susceptibility, phenotypic traits, and drug responses~\cite{snp_1, snp_2, snp_3}.

Therefore, we introduced a k-mer embedding approach that combines metagenomic context and string structure. In our methodology, contextual information refers to the relationships between k-mers situated closely within sequences, while structural information examines nucleotide patterns within a k-mer and their relations to other k-mers. Using this, we constructed a metagenomic graph that builds upon the De Bruijn Graph to capture not only the transitions of k-mers but also the structural similarities.

Given the advances in Graph Neural Networks (GNNs), e.g.~by~\citet{GCN}, we grounded our method in GNNs but designed for heterogeneous graphs. This approach effectively recognizes and uses both contextual and structural connection types.
Further, drawing from the success of self-supervised pre-training in Natural Language Processing and Computer Vision~\cite{bert, byol, SimCLR}, we designed a self-supervised objective for genomic graph data. We employed contrastive loss aiming to align closely in representation space k-mers with similar context and structure.

Finally, we benchmarked our technique on two downstream tasks: Edit Distance Approximation and Closest String Retrieval, influenced by~\citet{neuroseed}. The former estimates the minimum number of changes required to transform one genomic sequence into the other but without quadratic computational complexity. This is crucial, as understanding the evolutionary distance between constantly evolving sequences remains a primary challenge in biology. The latter task, Closest String Retrieval, involves efficiently finding sequences resembling a provided query, a method biologists use when categorizing new genes.
 
\section{Related Work}

\paragraph{Genomic Sequence Representation} Machine learning methods have emerged in computational biology to represent genomic sequences. A key component is the k-mer: a continuous nucleotide sequence of length \( k \). The Word2Vec method~\cite{word2vec}, which represents words as vectors using their context, treats overlapping k-mers in genomic sequences as words in sentences. Building on this, \mbox{\citet{kmer2vec}} introduced \emph{kmer2vec} to apply Word2Vec to genomic data for Multiple Sequence Alignment. Another strategy is to use the De Bruijn graph, where k-mers are nodes and their overlaps are edges, in conjunction with Node2Vec~\cite{node2vec}, which derives node features from the contextual information of biased random walks. This method underpins \citet{gradl}'s \emph{GRaDL} for early animal genome disease detection. K-mers also pair well with transformer-based models: \mbox{\citet{dnabert}'s} \emph{DNABERT} leverages a BERT-inspired objective~\cite{bert} and k-mer tokenization to predict genome-wide regulatory elements. Similarly, \citet{metagenome2vec}'s \emph{Metagenome2Vec} blends Node2Vec with transformers to analyze metagenomes with limited labeled data. Given the high computational demands of these transformer-based approaches, they fall outside the scope of our benchmarks in this study.

\paragraph{Graph Neural Networks} Graph Convolutional Networks (GCNs) are foundational to several innovative applications in graph-based machine learning~\cite{GCN}. In genomics, GNNs have been applied in metagenomic binning; for instance, \citet{metagenomic_binning_1}. As we aim to enhance our node embeddings with structural similarity, both heterogeneity and heterophily are key considerations.  Recognizing the ubiquity of heterogeneity in real-world graphs, Relational GCNs (R-GCNs) were developed. These networks expand upon GCNs by generalizing the convolution operation to handle different edge types, assigning distinct learnable weight matrices for each relation type~\cite{rgcn}. To tackle heterophily, where distant nodes in a graph may bear similar features, \citet{geom_gcn} introduced Geom-GCN, which maps nodes to a latent space, while \citet{beyond_homophily} suggested a distinct encoding approach for node embeddings and neighborhood aggregations.

\paragraph{Self-Supervised Learning} Self-supervised learning (SSL) enables effective use of unlabeled data and reduces dependence on annotated labels~\cite{ss_review}. Among SSL methods, contrastive learning has made a significant impact~\cite{byol, SimCLR}. At its core, contrastive learning seeks to bring similar data instances closer in the embedding space while pushing dissimilar ones apart. When applied to graph data, several techniques have been proposed for obtaining positive pairs, including uniform sampling, node dropping, and random walk sampling~\cite{ssgnn_review_10, ssgnn_review_49,ssgnn_review_48}.
 
\section{Methodology}

\subsection{Metagenomic Graph} \label{section:method:graph}

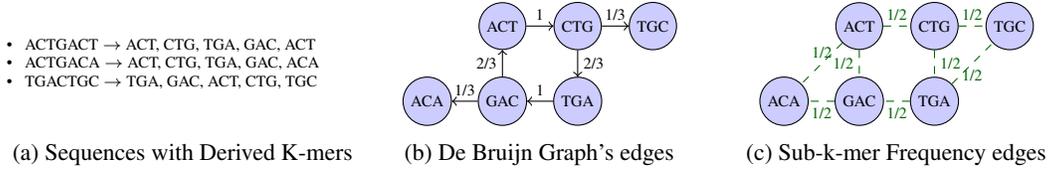
\begin{figure}[t]
    \centering
    \begin{subfigure}[b]{0.33\textwidth}
        \centering
            \tiny
             \begin{itemize}[leftmargin=1em, topsep=0pt, itemsep=0pt, parsep=0pt]
                \item ACTGACT $\rightarrow$ ACT, CTG, TGA, GAC, ACT
                \item ACTGACA $\rightarrow$ ACT, CTG, TGA, GAC, ACA
                \item TGACTGC $\rightarrow$ TGA, GAC, ACT, CTG, TGC
            \end{itemize}
            ~\\~
        \caption{Sequences with Derived K-mers}
        \label{fig:K-mers}
    \end{subfigure}~ \begin{subfigure}[b]{0.33\textwidth}
        \centering
        \begin{tikzpicture}[scale=0.5,auto,swap]
\tiny
            \node[circle,draw=black,fill=blue!20] (ACT) at (2,2) {ACT};
            \node[circle,draw=black,fill=blue!20] (CTG) at (4,2) {CTG};
            \node[circle,draw=black,fill=blue!20] (TGA) at (4,0) {TGA};
            \node[circle,draw=black,fill=blue!20] (GAC) at (2,0) {GAC};
            \node[circle,draw=black,fill=blue!20] (ACA) at (0,0) {ACA};
            \node[circle,draw=black,fill=blue!20] (TGC) at (6,2) {TGC};
            
\draw[->] (ACT) -- (CTG) node[midway,above] {1};
            \draw[->] (CTG) -- (TGA) node[midway,right] {2/3};
            \draw[->] (TGA) -- (GAC) node[midway,above] {1};
            \draw[->] (GAC) -- (ACT) node[midway,left] {2/3};
            \draw[->] (GAC) -- (ACA) node[midway,above] {1/3};
            \draw[->] (CTG) -- (TGC) node[midway,above] {1/3};
        \end{tikzpicture}
        \caption{\hyperlink{prg:db_edges}{De Bruijn Graph's edges}}
    \end{subfigure}~ \begin{subfigure}[b]{0.33\textwidth}
        \centering
        \begin{tikzpicture}[scale=0.5,auto,swap]
\tiny
            \node[circle,draw=black,fill=blue!20] (ACT) at (2,2) {ACT};
            \node[circle,draw=black,fill=blue!20] (CTG) at (4,2) {CTG};
            \node[circle,draw=black,fill=blue!20] (TGA) at (4,0) {TGA};
            \node[circle,draw=black,fill=blue!20] (GAC) at (2,0) {GAC};
            \node[circle,draw=black,fill=blue!20] (ACA) at (0,0) {ACA};
            \node[circle,draw=black,fill=blue!20] (TGC) at (6,2) {TGC};
            
\draw[color=black!60!green, dashed] (ACT) to[out=0,in=-180] node[midway,above] {1/2} (CTG);
            \draw[color=black!60!green, dashed] (ACT) to[out=-90,in=90] node[midway,left] {1/2} (GAC);
            \draw[color=black!60!green, dashed] (ACT) to[out=-135,in=45] node[midway,above] {1/2} (ACA);
            
            \draw[color=black!60!green, dashed] (CTG) to[out=-90,in=90] node[midway,right] {1/2} (TGA);
            \draw[color=black!60!green, dashed] (CTG) to[out=0,in=180] node[midway,above] {1/2} (TGC);
            
            \draw[color=black!60!green, dashed] (TGA) to[out=-180,in=0] node[midway,below] {1/2} (GAC);
            \draw[color=black!60!green, dashed] (TGA) to[out=45,in=-135] node[midway,below] {1/2} (TGC);
            
            \draw[color=black!60!green, dashed] (GAC) to[out=180,in=0] node[midway,below] {1/2} (ACA);
            
        \end{tikzpicture}
        \caption{\hyperlink{prg:kf_edges}{Sub-k-mer Frequency edges}}
        \label{fig:kf_graph_toy_example}
    \end{subfigure}
    \caption{Example of a Metagenomic Graph where nodes represent k-mers, De Bruijn Graph's edges capture contextual information, while Sub-k-mer Frequency edges depict structural similarity.}
    \label{fig:metagenomic_graph}
\end{figure}

The De Bruijn Graph, constructed from metagenomic sequences, forms the foundation of our method. In this graph, each k-mer, a substring of length \(k\) derived from the sequences, is uniquely represented by a different node. Additionally, an edge from node \( v_i \) to node \( v_j \) in the graph indicates that the k-mer at node \( v_i \) directly precedes the k-mer at node \( v_j \) in one of the sequences of the metagenome. When used, edge weights represent the frequency of these transitions, thereby capturing intrinsic genomic structures within the graph.

Although Node2Vec effectively captures the sequential context in De Bruijn graphs, it overlooks structural k-mer similarities. To address this, we expand the graph to include connections based on these similarities, complementing the transition probabilities. In subsequent sections, we detail the formulation of the two edge types for our graph, where nodes \(\{ v_i, v_j, \dots \}\) represent k-mers, as depicted in Figure~\ref{fig:metagenomic_graph}.

\hypertarget{prg:db_edges}{\paragraph{De Bruijn Graph's edges}} The first edge type is designed to capture contextual information.  Let \( T(v_i, v_j) \) represent the count of transitions between k-mers within a dataset of genomic sequences. The weight of an edge connecting nodes \( v_i \) and \( v_j \), \( w^{(dBG)}_{ij} \), is defined by,
\begin{equation}
    w^{(dBG)}_{ij} = \frac{T(v_i, v_j)}{\sum_{l \in  \delta^{+}(v_i)} T(v_i, v_l)},
\end{equation}
where \( \delta^{+}(v_i) \) denotes nodes adjacent to \( v_i \) via outgoing edges.

 \hypertarget{prg:kf_edges}{\paragraph{Sub-k-mer Frequency edges}} To efficiently capture the structural similarity between strings, we introduce a method using sub-k-mer frequency vectors, denoted as \( \mathbf{y}^{(\text{KF}_{\text{sub\_k}})}\). This vector quantifies the occurrences of each sub-k-mer of length sub\_k within a given k-mer. Specifically, the i-th entry indicates the frequency of the i-th sub-k-mer,
\begin{equation}
    \mathbf{y}^{(\text{KF}_{\text{sub\_k}})}[i] = \sum_{j=1}^{k - \text{sub\_k}+1} \mathbf{1}_{\text{k-mer}[j: j+\text{sub\_k} - 1] = s_i}, \forall i, s_i \in \Sigma^{\text{sub\_k}}.
\end{equation}
The k-mer similarity is then determined using the cosine similarity between these sub-k-mer frequency vectors,
\begin{equation}
    w_{ij}^{(\text{KF}_{\text{sub\_k}})} =
    \frac{
        {\mathbf{y}_i^{(\text{KF}_{\text{sub\_k}})}}^T
        \mathbf{y}_j^{(\text{KF}_{\text{sub\_k}})}}
        {\|\mathbf{y}_i^{(\text{KF}_{\text{sub\_k}})}\|_2 \|\mathbf{y}_j^{(\text{KF}_{\text{sub\_k}})}\|_2 }.
\end{equation}
This method, scaling linearly with the frequency vector size per weight, provides a computational advantage in practice over the direct Edit Distance calculation for k-mers, which exhibits \(k^2\) complexity per weight. Even so, computation for each node pair remains necessary. Given that node counts might approach \(4^k\), we apply edge-filtering at threshold \(t\), retaining only the links with the highest similarity. The filtered set of weights is then,
\begin{equation}
\mathbf{W^{\scriptstyle(\text{KF}_{\text{sub\_k}})}} = \{~ w^{\smash{(\text{KF}_{\text{sub\_k}})}}_{\scriptstyle ij} \mid w^{\smash{(\text{KF}_{\text{sub\_k}})}}_{\scriptstyle ij} \geq t ~\}.
\end{equation}

To better accommodate graphs for larger \(k\) values, we have also developed a more scalable approximation of the above approach. It utilizes state-of-the-art approximate nearest neighbor search \citep{faiss-johnson2019billion}  on the sub-k-mer frequency vectors, replacing the computationally demanding pairwise cosine similarity calculations. The details of this adaptation and its experimental results are outlined in Appendix~\ref{app:scalable_version}, demonstrating the method's effectiveness in processing large metagenomic graphs.

\paragraph{Notation}
The metagenomic graph is formally defined as \( G = (V, E, W) \).  In this graph, nodes \( V \) correspond to individual k-mers. The edges \( E \) can be categorized into two sets: De Bruijn Graphs's edges \( E^{(dBG)} \) and Sub-k-mer Frequency edges \( E^{(KF)} \). Edges in \( E^{(KF)} \) may be further subdivided based on various \( \text{sub\_k} \) values. Thus, edge weights \( W \) can contain \( \mathbf{W}^{(dBG)} \) and several \( \mathbf{W}^{(KF_{\text{sub\_k}})} \).

\subsection{Encoder}
We tailored GNNs for a heterogeneous metagenomic graph to capture nuanced k-mer relationships. The design employs varying depths of message passing: deeper for De Bruijn edges to capture broader context and shallower for similarity measures. Central to this GNN is the adapted Graph Convolutional Layer, formulated as:
\begin{equation}
    \mathbf{H}^{(l+1)} = \sigma \left( 
    \mathbf{\tilde{D}}^{(\text{edge\_type})^{-\frac{1}{2}}}
    \mathbf{\tilde{W}}^{(\text{edge\_type})}
    \mathbf{\tilde{D}}^{(\text{edge\_type})^{-\frac{1}{2}}} \mathbf{H}^{(l)} \mathbf{\Theta}^{(l)} \right),
\end{equation}
where \( \mathbf{\tilde{W}}^{(\text{edge\_type})} \) includes added self-loops and \(\mathbf{\tilde{D}}_{ii}\) is its diagonal degree matrix. The term \emph{edge\_type} refers to either \(dBG\) or \(KF_{\text{sub\_k}}\). The GCN layout consists of multiple layers, each characterized by a unique edge feature type and the number of channels.

\subsection{Self-Supervised Task}

\tikzstyle{block} = [rectangle, draw, fill=gray!20, 
    text width=6em, text centered, rounded corners, minimum height=2em]
\tikzstyle{line} = [draw, -latex']
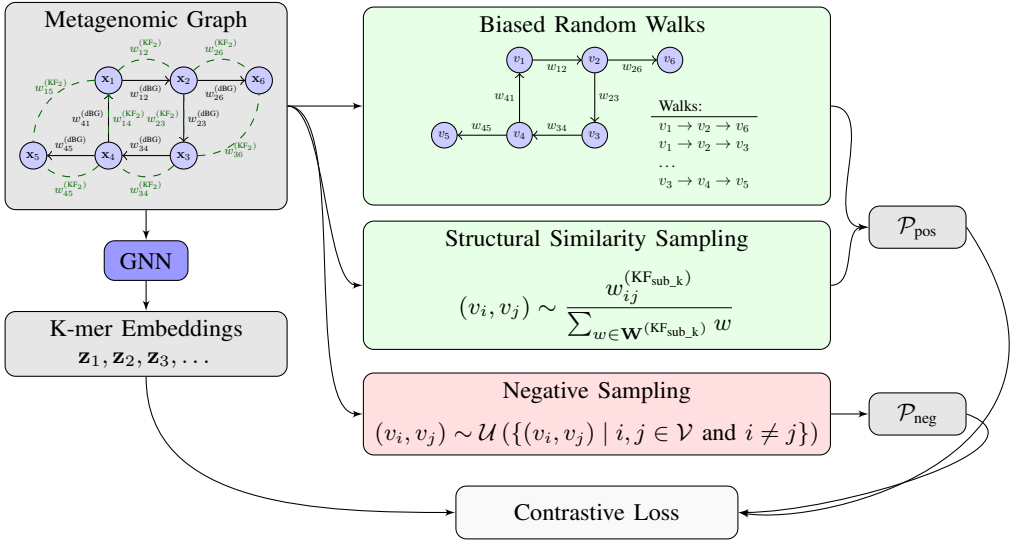
\begin{figure}[t]
    \centering
    
    \begin{tikzpicture}[auto, node distance = 3.5cm]

\node [block, text width=10em] (data) {\small Metagenomic Graph \\
         \begin{tikzpicture}[scale=0.5, transform shape]
            \node[circle,draw=black,fill=blue!20, minimum size=1mm] (ACT) at (2,2) {\normalsize  $\mathbf{x}_{1}$};
            \node[circle,draw=black,fill=blue!20, minimum size=1mm] (CTG) at (4,2) {\normalsize $\mathbf{x}_{2}$};
            \node[circle,draw=black,fill=blue!20, minimum size=1mm] (TGA) at (4,0) {\normalsize $\mathbf{x}_{3}$};
            \node[circle,draw=black,fill=blue!20, minimum size=1mm] (GAC) at (2,0) {\normalsize $\mathbf{x}_{4}$};
            \node[circle,draw=black,fill=blue!20, minimum size=1mm] (ACA) at (0,0) {\normalsize $\mathbf{x}_{5}$};
            \node[circle,draw=black,fill=blue!20, minimum size=1mm] (TGC) at (6,2) {\normalsize $\mathbf{x}_{6}$};

{\footnotesize
            \draw[->] (ACT) -- (CTG) node[midway,below] { $w^{(\text{dBG})}_{12}$};
            \draw[->] (CTG) -- (TGA) node[midway,right] { $w^{(\text{dBG})}_{23}$};
            \draw[->] (TGA) -- (GAC) node[midway,above] { $w^{(\text{dBG})}_{34}$};
            \draw[->] (GAC) -- (ACT) node[midway,left] { $w^{(\text{dBG})}_{41}$};
            \draw[->] (GAC) -- (ACA) node[midway,above] { $w^{(\text{dBG})}_{45}$};
            \draw[->] (CTG) -- (TGC) node[midway,below] { $w^{(\text{dBG})}_{26}$};

            \draw[color=black!60!green, dashed] (ACT) to[out=45,in=135] node[midway,above] { $w^{(\text{KF}_2)}_{12}$} (CTG);
            \draw[color=black!60!green, dashed] (ACT) to[out=-90,in=90] node[midway,right] { $w^{(\text{KF}_2)}_{14}$} (GAC);
            \draw[color=black!60!green, dashed] (ACT) to[out=180,in=90] node[midway,above] { $w^{(\text{KF}_2)}_{15}$} (ACA);
            
            \draw[color=black!60!green, dashed] (CTG) to[out=-90,in=90] node[midway,left] { $w^{(\text{KF}_2)}_{23}$} (TGA);
            \draw[color=black!60!green, dashed] (CTG) to[out=45,in=135] node[midway,above] { $w^{(\text{KF}_2)}_{26}$} (TGC);
            
            \draw[color=black!60!green, dashed] (TGA) to[out=-135,in=-45] node[midway,below] { $w^{(\text{KF}_2)}_{34}$} (GAC);
            \draw[color=black!60!green, dashed] (TGA) to[out=0,in=-90] node[midway,below] { 
            $w^{(\text{KF}_2)}_{36}$
            } (TGC);
            \draw[color=black!60!green, dashed] (GAC) to[out=-135,in=-45] node[midway,below] { $w^{(\text{KF}_2)}_{45}$} (ACA);
            }
                    
        \end{tikzpicture}
    
    };

    \node [block, fill=blue!40, below=0.4cm of data, text width=2.5em, minimum height=1.5em] (GNN) {\small GNN};

    \node [block, below=0.4cm of GNN, text width=10em] (embedding) {\small K-mer Embeddings \\
            $\mathbf{z}_{1}, \mathbf{z}_{2}, \mathbf{z}_{3}, \dots$ 
        
    };

    \node [block, fill=green!10, right= 1cm of data, text width=17em, minimum height=1em] (RW_sampling) {\small       Biased Random Walks \vspace{0.5em} \\
        \begin{tikzpicture}[scale=0.5, transform shape]
            \node[circle,draw=black,fill=blue!20, minimum size=1mm] (ACT) at (2,2) {\normalsize  $v_{1}$};
            \node[circle,draw=black,fill=blue!20, minimum size=1mm] (CTG) at (4,2) {\normalsize $v_{2}$};
            \node[circle,draw=black,fill=blue!20, minimum size=1mm] (TGA) at (4,0) {\normalsize $v_{3}$};
            \node[circle,draw=black,fill=blue!20, minimum size=1mm] (GAC) at (2,0) {\normalsize $v_{4}$};
            \node[circle,draw=black,fill=blue!20, minimum size=1mm] (ACA) at (0,0) {\normalsize $v_{5}$};
            \node[circle,draw=black,fill=blue!20, minimum size=1mm] (TGC) at (6,2) {\normalsize $v_{6}$};

\draw[->] (ACT) -- (CTG) node[midway,below] {\normalsize $w_{12}$};
            \draw[->] (CTG) -- (TGA) node[midway,right] {\normalsize $w_{23}$};
            \draw[->] (TGA) -- (GAC) node[midway,above] {\normalsize $w_{34}$};
            \draw[->] (GAC) -- (ACT) node[midway,left] {\normalsize $w_{41}$};
            \draw[->] (GAC) -- (ACA) node[midway,above] {\normalsize $w_{45}$};
            \draw[->] (CTG) -- (TGC) node[midway,below] {\normalsize $w_{26}$};

\large
            \node[right=1cm of TGA] (walks) {
                \begin{tabular}{l}
                    ~\\
                    Walks:\\
                    \hline
                    $v_1 \rightarrow v_2  \rightarrow v_6$ \\
                    $v_1 \rightarrow v_2 \rightarrow v_3$ \\
                    \dots \\
                    $v_3 \rightarrow v_4 \rightarrow v_5$ \\
                \end{tabular}
            };

        \end{tikzpicture}
    };

    \node [block, fill=green!10, below=0.2cm of RW_sampling, text width=17em, minimum height=1em] (KF_sampling) {\small Structural Similarity Sampling 
        $$
        (v_i, v_j) \sim \frac{w^{(\text{KF}_{\text{sub\_k}})}_{ij}}{\sum_{w \in \mathbf{W^{(\text{KF}_{\text{sub\_k}})}}} w}
        $$
        
    };

    \node [block, below right=0cm and 0.5cm of RW_sampling, text width=3em, minimum height=1.5em] (pos_samples) {\small
    \vspace{-1em}
    \[\mathcal{P}_{\text{pos}}\]
    };

    \node [block, fill=pink!50, below=0.3cm of KF_sampling, text width=17em, minimum height=1em] (negative_sampling) {\small Negative Sampling \[(v_i, v_j) \sim \mathcal{U}\left(\left\{ (v_i, v_j) \mid i, j \in \mathcal{V} \text{ and } i \neq j \right\}\right)\]};

    \node [block, right=0.5cm of negative_sampling, text width=3em, minimum height=1em] (neg_samples) {\small
    \vspace{-1em}
    \[\mathcal{P}_{\text{neg}}\]
    };

    \node [block, fill=gray!5, text width=10em, below=0.4cm of negative_sampling] (loss) {\small Contrastive Loss
    };

\path [line] (data) -- (GNN);
    \path [line] (GNN) -- (embedding);
    
    \path [line] (data.east) to[out=0,in=180, looseness=0.5] (RW_sampling.west);
    \path [line] (data.east) to[out=0,in=180, looseness=0.5] (KF_sampling.west);
    \path [line] (data.east) to[out=0,in=180, looseness=0.5] (negative_sampling.west);

    \path [line] (RW_sampling.east) to[out=0,in=180] (pos_samples.west);
    \path [line] (KF_sampling.east) to[out=0,in=180] (pos_samples.west);
    \path [line] (negative_sampling.east) to[out=0,in=180] (neg_samples.west);
    
    \path [line] (embedding.south) to[out=-90,in=180] (loss.west);
    \path [line] (pos_samples.east) .. controls +(2cm,-3cm) and +(6cm:-1.2cm) .. (loss.east);
    \path [line] (neg_samples.east) to[out=-20,in=0, looseness=1] (loss.east);

\end{tikzpicture}
    \caption{Self-Supervised Graph Contrastive Learning approach.}

    \label{fig:CL_workflow}
\end{figure}

We investigate the use of a contrastive learning method for k-mer representations. Graph nodes are initialized using a sub-k-mer frequency vector. Positive and negative pairs are sampled and, along with the k-mer representations from the encoder, are used to compute the loss, as depicted in Figure~\ref{fig:CL_workflow}.

\paragraph{Biased Random Walk Sampling} We employ Biased Random Walk Sampling to capture k-mer contextual information. This approach uses \(w^{(dBG)}\) edges to conduct walks, implemented exactly as in Node2Vec~\cite{node2vec}. Given a walk of a set length, we extract positive pairs by applying a window of size \(m\). Using a shrink factor \(\delta\), drawn uniformly from \(\{1, \ldots, m\}\), we determine the range \(i \pm \delta\) within which nodes are considered positive pairs to node \(v_i\). Repeating this across multiple random walks, we gather a comprehensive set of positive pairs.

\paragraph{Structural Similarity Sampling}  To capture the structural notion of k-mers, we sample pairs with probability proportional to sub-k-mer frequency similarity, \(w^{(\text{KF}_{\text{sub\_k}})}\). The aim is for k-mers linked by higher similarity (weights closer to 1) to have similar representations. The probability of sampling is given by,
\begin{equation}
    (v_i, v_j) \sim \frac{w^{(\text{KF}_{\text{sub\_k}})}_{ij}}{\sum_{w \in \mathbf{W^{(\text{KF}_{\text{sub\_k}})}}} w}.
\end{equation}
\paragraph{Negative Sampling} We randomly select negative pairs from all node pairs in the graph, leveraging the assumption that most such pairs lack a high similarity edge. This approach ensures diversity in learned representations.

\paragraph{Loss Function} Having established both positive \(\mathcal{P}_{\text{pos}}\) and negative \(\mathcal{P}_{\text{neg}}\) pair types, we apply the contrastive loss function. Following the approach by~\citet{graph_sage}, and using \( \sigma(x) \) as the sigmoid function, the equation is,
\begin{equation}
    l_{ij} = - \log \left( \sigma \left( \mathbf{z}_i^T \mathbf{z}_j \right) \right) - \sum_{(i,l) \in \mathcal{P}_{\text{neg}}(i)}\log \left( \sigma \left( -\mathbf{z}_i^T \mathbf{z}_l \right) \right).
\end{equation}
To reduce memory usage, we employed Neighborhood Sampling, again from~\cite{graph_sage}, for mini-batching during training.

\section{Bioinformatics Tasks}

\paragraph{Edit Distance Approximation} The task aims to calculate the edit distance without the burden of quadratic complexity. The \textit{NeuroSEED} framework by~\citet{neuroseed} offers a solution by providing sequence representations trained on a ground truth set of edit distances. In our experimental approach, we began with sequence representations derived from k-mer embeddings and subsequently fine-tuned them with a single linear layer. Our experiments were tested against One-Hot encoding (for \(k=1\) corresponding to \textit{NeuroSEED}~\cite{neuroseed}), Word2Vec, and Node2Vec. To find optimal hyperparameters, we executed a grid search on the validation set. Based on \citet{neuroseed} 's findings, we employed the hyperbolic function as it consistently outperformed other distance measures. Our primary metric for evaluation was the percentage Root Mean Squared Error (\% RMSE), where \( l \) denotes the dataset's maximum sequence length, \( h \) represents the hyperbolic distance function, and \( f_\theta \) indicates the downstream model,
\begin{equation}
    \% \text{RMSE}(\theta, \mathcal{D}) = \frac{100}{l} \sqrt{\sum_{s_1, s_2 \in \mathcal{D}} \left(\text{EditDistance}(s_1, s_2) - l \cdot h \left(f_\theta(\mathbf{\boldsymbol{z}_1}), f_\theta( \mathbf{\boldsymbol{z}_2})\right) \right)^2}.
\end{equation}

\paragraph{Closest String Retrieval}
The task is to find the sequence from a reference set that is closest to a specified query. We assessed embeddings fine-tuned on the edit distance approximation task using Convolutional Neural Networks (CNNs). These embeddings were contrasted with ones directly derived from our Self-supervised method, One-Hot, Word2Vec, or Node2Vec, through concatenation or taking the mean of k-mer embeddings. For performance assessment, we used top-\emph{n}\% accuracies, measuring how often the actual sequence appears within the top \emph{n}\% of positions based on the closeness of embedding vectors in hyperbolic space. We selected the optimal model for the embeddings based on the validation loss observed for the previous Edit Distance task. 
\section{Results and Analysis}

\begin{table}[!b]
\caption{RMSE \(\downarrow\) for the Edit Distance Approximation Task fine-tuned with Single Linear Layer. The best result per \(k\) is highlighted in bold, with deeper green shades indicating better performance across all runs. The standard deviation is based on three runs.}
\scriptsize
\centering
\setlength{\extrarowheight}{0pt}
\setlength{\tabcolsep}{3.9pt}
\addtolength{\extrarowheight}{\aboverulesep}
\addtolength{\extrarowheight}{\belowrulesep}
\setlength{\aboverulesep}{0pt}
\setlength{\belowrulesep}{0pt}

\begin{tabular}{r|cccc|cccc}
\toprule
& \multicolumn{4}{c|}{RT988 dataset} & \multicolumn{4}{c}{Qiita dataset} \\

$k$ & One-Hot & Word2Vec & Node2Vec & Our CL & One-Hot & Word2Vec & Node2Vec & Our CL \\ 
\midrule
1 & $\mathbf{0.43 \pm 0.01}$ & $0.44 \pm 0.01$ & $0.50 \pm 0.01$ & - & $\mathbf{2.46 \pm 0.03}$ & $2.57 \pm 0.01$ & $2.66 \pm 0.04$ & -~ \\
2 & $\mathbf{0.40 \pm 0.01}$ & $0.42 \pm 0.01$ & $0.45 \pm 0.01$ & $\mathbf{0.40 \pm 0.01}$ & $2.31 \pm 0.01$ & $2.22 \pm 0.03$ & $2.43 \pm 0.01$ & {\cellcolor[rgb]{0.988,1,0.988}}$\mathbf{2.14 \pm 0.02}$ \\
3 & $0.41 \pm 0.01$ & $0.42 \pm 0.01$ & $0.38 \pm 0.01$ & {\cellcolor[rgb]{0.98,1,0.98}}$\mathbf{0.37 \pm 0.01}$ & $2.41 \pm 0.01$ & $2.29 \pm 0.01$ & $2.29 \pm 0.04$ & {\cellcolor[rgb]{0.914,1,0.914}}$\mathbf{2.09 \pm 0.03}$ \\
4 & $0.42 \pm 0.01$ & $0.41 \pm 0.01$ & {\cellcolor[rgb]{0.996,1,0.996}}$0.38 \pm 0.01$ & {\cellcolor[rgb]{0.98,1,0.98}}$\mathbf{0.37 \pm 0.01}$ & $2.65 \pm 0.03$ & $2.35 \pm 0.01$ & $2.27 \pm 0.04$ & {\cellcolor[rgb]{0.784,1,0.784}}$\mathbf{2.00 \pm 0.01}$ \\
5 & $0.43 \pm 0.01$ & $0.40 \pm 0.01$ & {\cellcolor[rgb]{0.843,1,0.843}}$0.36 \pm 0.01$ & {\cellcolor[rgb]{0.502,1,0.502}}$\mathbf{0.35 \pm 0.01}$ & $3.14 \pm 0.01$ & $2.17 \pm 0.01$ & {\cellcolor[rgb]{0.996,1,0.996}}$2.16 \pm 0.02$ & {\cellcolor[rgb]{0.784,1,0.784}}$\mathbf{2.00 \pm 0.01}$ \\
6 & $0.43 \pm 0.01$ & $0.39 \pm 0.01$ & {\cellcolor[rgb]{0.918,1,0.918}}$0.37 \pm 0.01$ & {\cellcolor[rgb]{0.882,1,0.882}}$\mathbf{0.36 \pm 0.01}$ & $3.51 \pm 0.04$ & {\cellcolor[rgb]{0.784,1,0.784}}$2.00 \pm 0.01$ & {\cellcolor[rgb]{0.973,1,0.973}}$2.12 \pm 0.03$ & {\cellcolor[rgb]{0.502,1,0.502}}$\mathbf{1.97 \pm 0.01}$ \\
7 & $0.44 \pm 0.01$ & - & - & {\cellcolor[rgb]{0.882,1,0.882}}$\mathbf{0.36 \pm 0.01}$ & $4.12 \pm 0.07$ & -~ & -~ & {\cellcolor[rgb]{0.663,1,0.663}}$\mathbf{1.99 \pm 0.01}$ \\
8 & - & - & - & {\cellcolor[rgb]{0.502,1,0.502}}$\mathbf{0.35 \pm 0.01}$ & -~ & -~ & -~ & {\cellcolor[rgb]{0.502,1,0.502}}$\mathbf{1.96 \pm 0.01}$ \\
\bottomrule
\end{tabular}
\label{tbl:linear}
\end{table}

In all our experiments, the memory requirements of the One-Hot method increase exponentially, leading to its exclusion from our results for \(k > 7\). When pre-training exclusively on the training set, our method, thanks to the GCN encoder, can generalize beyond k-mers present in the training set. In contrast, Node2Vec and Word2Vec can only handle k-mer sizes up to the diversity of the training dataset. Hence, for $k>6$, where the test set introduces new k-mers, we had to exclude these methods.

\subsection{Edit Distance Approximation}
Table~\ref{tbl:linear} presents the results obtained using our pre-trained embeddings to estimate edit distances between sequences on the \emph{RT988} and \emph{Qiita} datasets from~\cite{neuroseed}. For the \emph{RT988} dataset, our Contrastive Learning (CL) and Node2Vec techniques surpass Word2Vec and One-Hot. The increased losses in \emph{Qiita} highlight its greater complexity. In this context, our method's integration of k-mer structural similarity becomes even more beneficial, outperforming all other tested methods. This benefit becomes more evident as \(k\) increases, underscoring our embedding's capability to adapt to new nodes.

\subsection{Closest String Retrieval}
\begin{table}[!b]
\centering
\caption{Mean Top retrieval performance. Best results per \(k\) are highlighted in bold, with deeper green shades indicating better performance separately for Zero-Shot and Fine-Tuned. The standard deviation is based on three runs.}
\label{tbl:top}
\begin{subtable}{\textwidth}
\caption{Top 1\% $\uparrow$}
\label{tbl:top1}
\scriptsize
\centering
\setlength{\extrarowheight}{0pt}
\addtolength{\extrarowheight}{\aboverulesep}
\addtolength{\extrarowheight}{\belowrulesep}
\setlength{\aboverulesep}{0pt}
\setlength{\belowrulesep}{0pt}

\begin{tabular}{r|cccc|cccc} 
\toprule
& \multicolumn{4}{c|}{Zero-Shot: Aggregated K-mer Embeddings} & \multicolumn{4}{c}{\begin{tabular}[c]{@{}c@{}}Fine-Tuned: \textit{NeuroSEED} with K-mer Embeddings\\(equivalent to \citet{neuroseed} for One-Hot at $k=1$)\end{tabular}} \\
\cline{2-9}
$k$ & One-Hot & Word2Vec & Node2Vec & Our CL & One-Hot & Word2Vec & Node2Vec & Our CL \\
\hline
1 & {\cellcolor[rgb]{0.867,1,0.867}}$\mathbf{50.3}$ & {\cellcolor[rgb]{0.988,1,0.988}}$45 \pm 0.1$ & {\cellcolor[rgb]{0.984,1,0.984}}$45.3 \pm 1.5$ & - & {\cellcolor[rgb]{0.973,1,0.973}}$46.9 \pm 0.9$ & {\cellcolor[rgb]{0.957,1,0.957}}$47.5 \pm 1.3$ & {\cellcolor[rgb]{0.992,1,0.992}}$46.2 \pm 0.1$ & - \\
2 & {\cellcolor[rgb]{0.914,1,0.914}}$49.9$ & {\cellcolor[rgb]{0.969,1,0.969}}$46.7 \pm 0.1$ & {\cellcolor[rgb]{0.914,1,0.914}}$49.9 \pm 0.6$ & {\cellcolor[rgb]{0.502,1,0.502}}$\mathbf{52.2 \pm 0.7}$ & {\cellcolor[rgb]{0.914,1,0.914}}$48 \pm 0.1$ & {\cellcolor[rgb]{0.933,1,0.933}}$47.8 \pm 1.8$ & {\cellcolor[rgb]{0.949,1,0.949}}$47.6 \pm 0.5$ & {\cellcolor[rgb]{0.89,1,0.89}}${48.3 \pm 0.7}$ \\
3 & {\cellcolor[rgb]{0.965,1,0.965}}$46.8$ & {\cellcolor[rgb]{0.945,1,0.945}}$48.6 \pm 0.1$ & {\cellcolor[rgb]{0.502,1,0.502}}$51.2 \pm 0.2$ & {\cellcolor[rgb]{0.502,1,0.502}}$\mathbf{53.1 \pm 0.4}$ & {\cellcolor[rgb]{0.984,1,0.984}}$46.4 \pm 1$ & {\cellcolor[rgb]{0.961,1,0.961}}$47.3 \pm 0.1$ & {\cellcolor[rgb]{0.859,1,0.859}}${49.1 \pm 0.6}$ & {\cellcolor[rgb]{0.902,1,0.902}}$48.2 \pm 0.3$ \\
4 & {\cellcolor[rgb]{0.984,1,0.984}}$45.3$ & {\cellcolor[rgb]{0.961,1,0.961}}$46.9 \pm 0.1$ & {\cellcolor[rgb]{0.925,1,0.925}}$49.8 \pm 0.2$ & {\cellcolor[rgb]{0.502,1,0.502}}$\mathbf{53.3 \pm 0.3}$ & {\cellcolor[rgb]{0.992,1,0.992}}$46.2 \pm 0.5$ & {\cellcolor[rgb]{0.976,1,0.976}}$46.8 \pm 1.5$ & {\cellcolor[rgb]{0.902,1,0.902}}${48.2 \pm 1.3}$ & {\cellcolor[rgb]{0.941,1,0.941}}$47.7 \pm 0.8$ \\
5 & {\cellcolor[rgb]{0.976,1,0.976}}$45.4$ & {\cellcolor[rgb]{0.996,1,0.996}}$42.3 \pm 0.1$ & {\cellcolor[rgb]{0.898,1,0.898}}$50 \pm 0.4$ & {\cellcolor[rgb]{0.859,1,0.859}}$\mathbf{50.5 \pm 0.1}$ & {\cellcolor[rgb]{0.98,1,0.98}}$46.6 \pm 0.8$ & {\cellcolor[rgb]{0.969,1,0.969}}$47 \pm 1.1$ & {\cellcolor[rgb]{0.871,1,0.871}}${48.8 \pm 1.8}$ & {\cellcolor[rgb]{0.984,1,0.984}}$47.8 \pm 0.3$ \\
6   & {\cellcolor[rgb]{0.973,1,0.973}}$46.3$ & $41.3 \pm 0.1$ & {\cellcolor[rgb]{0.933,1,0.933}}$49.6 \pm 0.3$ & {\cellcolor[rgb]{0.898,1,0.898}}${50 \pm 0.7}$ & {\cellcolor[rgb]{0.89,1,0.89}}$48.3 \pm 1.2$ & {\cellcolor[rgb]{0.996,1,0.996}}$45.2 \pm 0.5$ & {\cellcolor[rgb]{0.502,1,0.502}}$\mathbf{50.1 \pm 0.4}$ & {\cellcolor[rgb]{0.969,1,0.969}}$47 \pm 0.9$ \\
7   & {\cellcolor[rgb]{0.992,1,0.992}}$44.5$ & - & - & {\cellcolor[rgb]{0.949,1,0.949}}${48.3 \pm 1.1}$ & {\cellcolor[rgb]{0.996,1,0.996}}$44.8 \pm 0.5$ & - & - & {\cellcolor[rgb]{0.867,1,0.867}}$\mathbf{48.9 \pm 0.6}$ \\
8   & - & - & - & {\cellcolor[rgb]{0.878,1,0.878}}$\mathbf{50.2 \pm 0.1}$ & - & - & - & {\cellcolor[rgb]{0.914,1,0.914}}${48 \pm 0.3}$ \\
\bottomrule
\end{tabular}
\end{subtable}
~
\begin{subtable}{\textwidth}
\caption{Top 10\% $\uparrow$}
\footnotesize
\label{tbl:top10}
\scriptsize
\centering
\setlength{\extrarowheight}{0pt}
\addtolength{\extrarowheight}{\aboverulesep}
\addtolength{\extrarowheight}{\belowrulesep}
\setlength{\aboverulesep}{0pt}
\setlength{\belowrulesep}{0pt}
\begin{tabular}{r|cccc|cccc} 
\toprule
& \multicolumn{4}{c|}{Zero-Shot: Aggregated K-mer Embeddings} & \multicolumn{4}{c}{\begin{tabular}[c]{@{}c@{}}Fine-Tuned: \textit{NeuroSEED} with K-mer Embeddings\\(equivalent to \citet{neuroseed} for One-Hot at $k=1$)\end{tabular}
} \\
\cline{2-9}
$k$ & One-Hot & Word2Vec & Node2Vec & Our CL & One-Hot & Word2Vec & Node2Vec & Our CL \\ 
\hline
1   & {\cellcolor[rgb]{0.996,1,0.996}}$60.1$ & $58.0 \pm 0.1$ & {\cellcolor[rgb]{0.992,1,0.992}}${60.5 \pm 0.3}$ & - & {\cellcolor[rgb]{0.871,1,0.871}}$\mathbf{75.9 \pm 1.2}$ & {\cellcolor[rgb]{0.969,1,0.969}}$75.2 \pm 0.9$ & {\cellcolor[rgb]{0.988,1,0.988}}$74.9 \pm 0.7$ & - \\
2   & {\cellcolor[rgb]{0.988,1,0.988}}$60.7$ & {\cellcolor[rgb]{0.992,1,0.992}}$60.3 \pm 0.1$ & {\cellcolor[rgb]{0.996,1,0.996}}$60.1 \pm 0.4$ & {\cellcolor[rgb]{0.941,1,0.941}}${68.0 \pm 1.1}$ & {\cellcolor[rgb]{0.922,1,0.922}}$75.4 \pm 0.1$ & {\cellcolor[rgb]{0.863,1,0.863}}$76 \pm 0.8$ & {\cellcolor[rgb]{0.949,1,0.949}}$75.3 \pm 0.7$ & {\cellcolor[rgb]{0.851,1,0.851}}$\mathbf{76.4 \pm 0.5}$ \\
3   & {\cellcolor[rgb]{0.98,1,0.98}}$61.4$ & {\cellcolor[rgb]{0.984,1,0.984}}$61.3 \pm 0.1$ & {\cellcolor[rgb]{0.961,1,0.961}}$64.9 \pm 0.2$ & {\cellcolor[rgb]{0.914,1,0.914}}${70.8 \pm 0.6}$ & {\cellcolor[rgb]{0.969,1,0.969}}$75.2 \pm 0.4$ & {\cellcolor[rgb]{0.984,1,0.984}}$75 \pm 0.6$ & {\cellcolor[rgb]{0.922,1,0.922}}$75.4 \pm 0.3$ & {\cellcolor[rgb]{0.89,1,0.89}}$\mathbf{75.6 \pm 0.6}$ \\
4   & {\cellcolor[rgb]{0.965,1,0.965}}$64.6$ & {\cellcolor[rgb]{0.969,1,0.969}}$64.4 \pm 0.3$ & {\cellcolor[rgb]{0.902,1,0.902}}$73.8 \pm 0.3$ & {\cellcolor[rgb]{0.878,1,0.878}}$\mathbf{78.1 \pm 0.1}$ & {\cellcolor[rgb]{0.992,1,0.992}}$74.6 \pm 0.1$ & {\cellcolor[rgb]{0.969,1,0.969}}$75.2 \pm 0.5$ & {\cellcolor[rgb]{0.976,1,0.976}}$75.1 \pm 1.6$ & {\cellcolor[rgb]{0.922,1,0.922}}${75.4 \pm 0.3}$ \\
5   & {\cellcolor[rgb]{0.925,1,0.925}}$68.5$ & {\cellcolor[rgb]{0.973,1,0.973}}$62.7 \pm 0.2$ & {\cellcolor[rgb]{0.898,1,0.898}}$74.3 \pm 0.2$ & {\cellcolor[rgb]{0.867,1,0.867}}$\mathbf{79.5 \pm 0.2}$ & {\cellcolor[rgb]{0.878,1,0.878}}$75.8 \pm 0.3$ & {\cellcolor[rgb]{0.949,1,0.949}}$75.3 \pm 0.8$ & {\cellcolor[rgb]{0.863,1,0.863}}${76 \pm 1.9}$ & {\cellcolor[rgb]{0.949,1,0.949}}$75.3 \pm 0.8$ \\
6   & {\cellcolor[rgb]{0.933,1,0.933}}$68.4$ & {\cellcolor[rgb]{0.953,1,0.953}}$67.3 \pm 0.1$ & {\cellcolor[rgb]{0.91,1,0.91}}$71.4 \pm 0.3$            & {\cellcolor[rgb]{0.502,1,0.502}}$\mathbf{81.3 \pm 0.1}$  & {\cellcolor[rgb]{0.89,1,0.89}}$75.6 \pm 0.1$            & {\cellcolor[rgb]{0.996,1,0.996}}$74.1 \pm 0.9$ & {\cellcolor[rgb]{0.502,1,0.502}}${78.1 \pm 1.0}$ & {\cellcolor[rgb]{0.992,1,0.992}}$74.8 \pm 0.7$           \\
7  & $65.6$                                 & -                                              & -                                                       & {\cellcolor[rgb]{0.855,1,0.855}}$\mathbf{80.1 \pm 0.5}$  & $71.2 \pm 0.3$                                          & -                                              & -                                                       & {\cellcolor[rgb]{0.502,1,0.502}}${77.8 \pm 0.5}$  \\
8   & -                                      & -                                              & -                                                       & {\cellcolor[rgb]{0.867,1,0.867}}$\mathbf{79.4 \pm 0.2}$ & -                                                       & -                                              & -                                                       & {\cellcolor[rgb]{0.922,1,0.922}}${75.4 \pm 0.4}$  \\
\bottomrule
\end{tabular}
\end{subtable}
\end{table}

Tables \ref{tbl:top1} and \ref{tbl:top10} present the performance of our zero-shot sequence embeddings, directly derived from the aggregation of our k-mer embeddings, in retrieving the nearest sequences in the \emph{Qiita} dataset from~\cite{neuroseed}. The tables also showcase a comparison with the embeddings that were specifically fine-tuned for the Edit Distance Task, a process outlined by \citet{neuroseed}. 

For direct k-mer aggregation, our Contrastive Learning (CL) embeddings are obtained through concatenation, while for k-mer aggregation with One-Hot, Word2Vec, and Node2Vec, we report the results of the better performing method, either concatenation or averaging. The superior zero-shot non-parametric retrieval performance of our CL method emphasizes the combined utility of both context and structural similarity during self-supervised pre-training. Notably, while k-mers of size around three are optimal for Top 1\% retrieval, larger k-mers excel in the Top 10\% metrics. This suggests that smaller k-mers are better at discerning local sequence distances, whereas larger ones capture broader sequence distances.

For embeddings fine-tuned using CNNs for Edit Distance Approximation, the complexity of CNNs appears to obscure differences between the embeddings. Notably, our method based solely on zero-shot concatenated k-mer embeddings outperforms this complex fine-tuning. This shows the clear advantage of our embeddings over the \emph{NeuroSEED} method by~\citet{neuroseed}.
 
\section{Conclusion}

In our study, we introduced a novel k-mer embedding technique that seamlessly integrates metagenomic contextual and structural nuances, achieved through the enhancement of the De Bruijn graph and the use of contrastive learning. In the Edit Distance Approximation task, our technique consistently demonstrated superior performance compared to One-Hot, Word2Vec, and Node2Vec. Moreover, without requiring any downstream fine-tuning, our aggregated k-mer embeddings outperformed the \emph{Neuroseed} method by \citet{neuroseed} in the Closest String Retrieval task. These findings suggest potential broader uses in computational biology. 
\section{Acknowledgements}
Amir Joudaki is funded through Swiss National Science Foundation Project Grant \#200550 to Andre Kahles, and partially funded by ETH Core funding award to Gunnar Ratsch. Manuel Burger is funded by grant \#2022-278 of the Strategic Focus Area "Personalized Health and Related Technologies (PHRT)" of the ETH Domain (Swiss Federal Institutes of Technology).

\printbibliography

\newpage
\appendix

\section{Scalable K-mer Graph Construction}
\label{app:scalable_version}

This appendix addresses the challenge of assembling k-mer graphs for larger \(k\) values, offering a method more efficient than the pairwise cosine similarity calculations in our original framework. We incorporate the FAISS library \citep{faiss-johnson2019billion}, which uses an inverted file structure for efficient approximate nearest neighbor searches. This library identifies a  predetermined number of nearest neighbors for each node in the metagenomic graph, forming Sub-k-mer Frequency edges weighted according to distance metrics from the search. The process can be used for more than one sub-k value, potentially generating several subtypes of edges.

The effectiveness of this extension is demonstrated in edit distance approximation and closest string retrieval tasks, as presented in Tables \ref{tbl:faiss_linear} and \ref{tbl:faiss_top}. Performance for larger k-mers (\(k = 10\) and \(k = 15\)) is consistent with that of smaller k-mers, matching or surpassing Node2Vec and Word2Vec benchmarks. Unlike these benchmarks, our method effectively generates embeddings for new k-mers, unseen in the training data, thus facilitating scalability for larger \(k\). However, performance declines for very large k-mers (\(k = 20\) and \(k = 30\)). This decline is likely due to the high uniqueness of k-mers at these sizes in our datasets, reducing the informativeness of transition probabilities and the relevance of graph-based structural similarities.

\begin{table}[!h]
\caption{RMSE \(\downarrow\) for the Edit Distance Approximation Task on larger \(k\), fine-tuned with a single linear layer. Results derived using our contrastive learning framework with approximate nearest neighbor search instead of cosine similarity. The standard deviation is based on three runs.}
\scriptsize
\centering
\setlength{\extrarowheight}{0pt}
\setlength{\tabcolsep}{3.9pt}
\addtolength{\extrarowheight}{\aboverulesep}
\addtolength{\extrarowheight}{\belowrulesep}
\setlength{\aboverulesep}{0pt}
\setlength{\belowrulesep}{0pt}

\begin{tblr}{
  row{1} = {c},
  row{2} = {c},
  cell{1}{1} = {r},
  cell{2}{1} = {r},
  cell{3}{2} = {c},
  cell{3}{3} = {c},
  cell{4}{2} = {c},
  cell{4}{3} = {c},
  cell{5}{2} = {c},
  cell{5}{3} = {c},
  cell{6}{2} = {c},
  cell{6}{3} = {c},
  vline{2-3} = {-}{},
  hline{1,7} = {-}{0.08em},
  hline{3} = {-}{0.05em},
}
     & RT988 dataset   & Qiita dataset   \\
$k$  & Our CL          & Our CL          \\
$10$ & $0.36\pm 0.01$  & $2.05 \pm 0.01$ \\
$15$ & $0.36 \pm 0.01$ & $2.01\pm 0.02$  \\
$20$ & $0.36 \pm 0.01$ & $2.12\pm 0.01$  \\
$30$ & $0.37 \pm 0.01$ & $2.36 \pm 0.02$ 
\end{tblr}
\label{tbl:faiss_linear}
\end{table}

\begin{table}[!h]
\centering
\caption{Mean Top retrieval performance on larger \(k\). Results derived using our contrastive learning framework with approximate nearest neighbor search instead of cosine similarity. The standard deviation is based on three runs.}
\label{tbl:faiss_top}
\begin{subtable}{\textwidth}
\caption{Top 1\% $\uparrow$}
\scriptsize
\centering
\setlength{\extrarowheight}{0pt}
\addtolength{\extrarowheight}{\aboverulesep}
\addtolength{\extrarowheight}{\belowrulesep}
\setlength{\aboverulesep}{0pt}
\setlength{\belowrulesep}{0pt}

\begin{tabular}{l|c|c} 
\toprule
\multicolumn{1}{r|}{}                          & Zero-Shot: Aggregated K-mer Embeddings & Fine-Tuned: \textit{NeuroSEED} with K-mer Embeddings  \\ 
\cline{2-3}
\multicolumn{1}{r|}{$k$}                       & Our CL                                 & Our CL                                                \\ 
\hline
\begin{tabular}[c]{@{}l@{}}$10$\\\end{tabular} & $49.3 \pm 0.4$                         & $46.5 \pm 0.4$                                        \\
$15$                                           & $48.7 \pm 0.6$                         & $46.0 \pm 0.6$                                        \\
$20$                                           & $44.0 \pm 0.8$                         & $43.8 \pm 0.5$                                        \\
$30$                                           & $41.1\pm 0.9$                          & $42.1 \pm 0.6$                                        \\
\bottomrule
\end{tabular}
\end{subtable}
~
\begin{subtable}{\textwidth}
\caption{Top 10\% $\uparrow$}
\scriptsize
\centering
\setlength{\extrarowheight}{0pt}
\addtolength{\extrarowheight}{\aboverulesep}
\addtolength{\extrarowheight}{\belowrulesep}
\setlength{\aboverulesep}{0pt}
\setlength{\belowrulesep}{0pt}
\begin{tabular}{l|c|c} 
\toprule
\multicolumn{1}{r|}{}                          & Zero-Shot: Aggregated K-mer Embeddings & Fine-Tuned: \textit{NeuroSEED} with K-mer Embeddings  \\ 
\cline{2-3}
\multicolumn{1}{r|}{$k$}                       & Our CL                                 & Our CL                                                \\ 
\hline
\begin{tabular}[c]{@{}l@{}}$10$\\\end{tabular} & $77.6 \pm 1.2$                         & $71.9 \pm 0.5$                                        \\
$15$                                           & $78.4 \pm 0.8$                         & $71.9 \pm 0.4$                                        \\
$20$                                           & $76.1 \pm 0.8$                         & $70.8 \pm 0.4$                                        \\
$30$                                           & $71.3\pm 0.9$                          & $69.1 \pm 0.9$                                        \\
\bottomrule
\end{tabular}
\end{subtable}
\end{table}

\newpage
\section{Analysis of Sampling Techniques in Contrastive Learning}

Table~\ref{tbl:loss_analysis} presents the impact of different graph edges and corresponding sampling methods on the Edit Distance Approximation task. The results are presented for three distinct scenarios: training exclusively with dBG edges using Biased Random Walk Sampling, training exclusively with KF edges using Structural Similarity Sampling, and the standard approach that combines both edge types. These results highlight that integrating contextual and structural knowledge yields superior performance compared to employing each sampling strategy separately.

\begin{table}[!h]
\scriptsize
\centering
\caption{RMSE \(\downarrow\) for the Edit Distance Approximation Task. The standard deviation is based on three runs.}
\label{tbl:loss_analysis}
\begin{tblr}{
  row{odd} = {c},
  row{4} = {c},
  column{1} = {r},
  cell{1}{2} = {c=3}{},
  cell{2}{2} = {c},
  vline{2} = {-}{},
  hline{1,6} = {-}{0.08em},
  hline{3} = {-}{},
}
    & Qiita dataset   &                 &                          \\
$k$ & dBG edges only  & KF edges only   & Both Edge Types          \\
$3$ & $2.14 \pm 0.01$ & $2.12 \pm 0.02$ & $\mathbf{2.09 \pm 0.03}$ \\
$6$ & $2.19 \pm 0.03$ & $2.13 \pm 0.01$ & $\mathbf{1.97 \pm 0.01}$ \\
$8$ & $2.04 \pm 0.01$ & $2.09 \pm 0.01$ & $\mathbf{1.96 \pm 0.01}$ 
\end{tblr}
\end{table}

\section{Analysis of Graph Autoencoder as an Alternative Self-Supervised Task to Contrastive Learning}
This appendix outlines an alternative self-supervised learning task that applies a Graph Autoencoder (GAE) to the same Metagenomic Graph as the original method. This task eliminates the need for sampling, potentially offering computational benefits, especially for large graphs.

Figure~\ref{fig:GAE_workflow} illustrates the GAE methodology. After the encoding stage, our model employs two types of decoders: an edge decoder and a node decoder. Both are designed with simplicity in mind to avoid overfitting.

\tikzstyle{block} = [rectangle, draw, fill=gray!20, 
    text width=6em, text centered, rounded corners, minimum height=2em]
\tikzstyle{line} = [draw, -latex']
\begin{figure}[!h]
    \centering
    
    \begin{tikzpicture}[auto, node distance = 3.5cm]

\node [block, text width=10em] (data) {\small Metagenomic Graph \\
         \begin{tikzpicture}[scale=0.5, transform shape]
            \node[circle,draw=black,fill=blue!20, minimum size=1mm] (ACT) at (2,2) {\normalsize  $\mathbf{x}_{1}$};
            \node[circle,draw=black,fill=blue!20, minimum size=1mm] (CTG) at (4,2) {\normalsize $\mathbf{x}_{2}$};
            \node[circle,draw=black,fill=blue!20, minimum size=1mm] (TGA) at (4,0) {\normalsize $\mathbf{x}_{3}$};
            \node[circle,draw=black,fill=blue!20, minimum size=1mm] (GAC) at (2,0) {\normalsize $\mathbf{x}_{4}$};
            \node[circle,draw=black,fill=blue!20, minimum size=1mm] (ACA) at (0,0) {\normalsize $\mathbf{x}_{5}$};
            \node[circle,draw=black,fill=blue!20, minimum size=1mm] (TGC) at (6,2) {\normalsize $\mathbf{x}_{6}$};

{\footnotesize
            \draw[->] (ACT) -- (CTG) node[midway,below] { $w^{(\text{dBG})}_{12}$};
            \draw[->] (CTG) -- (TGA) node[midway,right] { $w^{(\text{dBG})}_{23}$};
            \draw[->] (TGA) -- (GAC) node[midway,above] { $w^{(\text{dBG})}_{34}$};
            \draw[->] (GAC) -- (ACT) node[midway,left] { $w^{(\text{dBG})}_{41}$};
            \draw[->] (GAC) -- (ACA) node[midway,above] { $w^{(\text{dBG})}_{45}$};
            \draw[->] (CTG) -- (TGC) node[midway,below] { $w^{(\text{dBG})}_{26}$};

            \draw[color=black!60!green, dashed] (ACT) to[out=45,in=135] node[midway,above] { $w^{(\text{KF}_2)}_{12}$} (CTG);
            \draw[color=black!60!green, dashed] (ACT) to[out=-90,in=90] node[midway,right] { $w^{(\text{KF}_2)}_{14}$} (GAC);
            \draw[color=black!60!green, dashed] (ACT) to[out=180,in=90] node[midway,above] { $w^{(\text{KF}_2)}_{15}$} (ACA);
            
            \draw[color=black!60!green, dashed] (CTG) to[out=-90,in=90] node[midway,left] { $w^{(\text{KF}_2)}_{23}$} (TGA);
            \draw[color=black!60!green, dashed] (CTG) to[out=45,in=135] node[midway,above] { $w^{(\text{KF}_2)}_{26}$} (TGC);
            
            \draw[color=black!60!green, dashed] (TGA) to[out=-135,in=-45] node[midway,below] { $w^{(\text{KF}_2)}_{34}$} (GAC);
            \draw[color=black!60!green, dashed] (TGA) to[out=0,in=-90] node[midway,below] { 
            $w^{(\text{KF}_2)}_{36}$
            } (TGC);
            \draw[color=black!60!green, dashed] (GAC) to[out=-135,in=-45] node[midway,below] { $w^{(\text{KF}_2)}_{45}$} (ACA);
            }
                    
        \end{tikzpicture}
    
    };

    \node [block, fill=blue!40, right=6em of data, text width=2.5em, minimum height=1.5em] (GNN) {\small GNN};

    \node [block, right of=GNN, text width=8em] (embedding) {\small K-mer Embeddings \\
            $\mathbf{z}_{1}, \mathbf{z}_{2}, \mathbf{z}_{3}, \dots$ 
    };

    \node [block, fill=blue!30, text width=3cm, below of=embedding, text width=12em] (n_decoder_1) {\small Node Decoder 1
        $
        \mathbf{\hat{y}_i^{(\text{KF}_2)}} 
            = 
        \mathbf{\Theta^{(\text{KF}_2)}} \mathbf{z}_i 
            + 
        \mathbf{b^{(\text{KF}_2)}}
        $
    };

\node [block, text width=9em, left=1em of n_decoder_1] (node_d_output) {
         \begin{tikzpicture}[scale=0.5, transform shape]
            \node[circle,draw=black,fill=blue!20, minimum size=1mm] (ACT) at (1,1) {\small $\mathbf{\hat{y}^{\text{KF}_2}_{1}}$};
            \node[circle,draw=black,fill=blue!20, minimum size=1mm] (CTG) at (3,1) {$\mathbf{\hat{y}^{\text{KF}_2}_{2}}$};
            \node[circle,draw=black,fill=blue!20, minimum size=1mm] (TGA) at (4,0) {$\mathbf{\hat{y}^{\text{KF}_2}_{3}}$};
            \node[circle,draw=black,fill=blue!20, minimum size=1mm] (GAC) at (2,0) {$\mathbf{\hat{y}^{\text{KF}_2}_{4}}$};
            \node[circle,draw=black,fill=blue!20, minimum size=1mm] (ACA) at (0,0) {$\mathbf{\hat{y}^{\text{KF}_2}_{5}}$};
            \node[circle,draw=black,fill=blue!20, minimum size=1mm] (TGC) at (5,1) {$\mathbf{\hat{y}^{\text{KF}_2}_{6}}$};                    
        \end{tikzpicture}
    
    };

    \node [block, fill=blue!30, text width=3cm, above= 2.5em of  n_decoder_1, node distance = 1.5cm, text width=12em] (edge_decoder) {\small Edge Decoder\\
        $\hat{w}^{(\text{dBG})}_{ij} = \mathbf{z}_i^\top \mathbf{z}_j$
    };

\node [block, text width=9em, left=1em of edge_decoder] (inner_d_output) {\normalsize
         \begin{tikzpicture}[scale=0.7, transform shape]
            \node[circle,draw=black,fill=blue!20, minimum size=1mm] (ACT) at (1,1) {};
            \node[circle,draw=black,fill=blue!20, minimum size=1mm] (CTG) at (2,1) {};
            \node[circle,draw=black,fill=blue!20, minimum size=1mm] (TGA) at (2,0) {};
            \node[circle,draw=black,fill=blue!20, minimum size=1mm] (GAC) at (1,0) {};
            \node[circle,draw=black,fill=blue!20, minimum size=1mm] (ACA) at (0,0) {};
            \node[circle,draw=black,fill=blue!20, minimum size=1mm] (TGC) at (3,1) {};

\draw[->] (ACT) -- (CTG) node[midway,above] { $\hat{w}^{(\text{dBG})}_{1,2}$};
            \draw[->] (CTG) -- (TGA) node[midway,right] { $\hat{w}^{(\text{dBG})}_{2,3}$};
            \draw[->] (TGA) -- (GAC) node[midway,below] { $\hat{w}^{(\text{dBG})}_{3,4}$};
            \draw[->] (GAC) -- (ACT) node[midway,left] {  $\hat{w}^{(\text{dBG})}_{4,1}$};
            \draw[->] (GAC) -- (ACA) node[midway,below] {  $\hat{w}^{(\text{dBG})}_{4,5}$};
            \draw[->] (CTG) -- (TGC) node[midway,above] { $\hat{w}^{(\text{dBG})}_{2,6}$};
                    
        \end{tikzpicture}
    
    };

    \node [block, text width=12em, below of= n_decoder_1, node distance = 1.5cm, fill=none, draw=none] (n_decoder_more) {\small     \dots 
    };

\node [block, text width=9em, left=1em of n_decoder_more,  fill=none, draw=none] (node_more_d_output) {
         \small \dots
    };

    \node [block, fill=gray!5, text width=6em, left of=node_d_output, node distance = 4.5cm] (losses) {\small Reconstruction Loss
    };

\path [line] (data) -- (GNN);
    \path [line] (GNN) -- (embedding);
    \path [line] (embedding.east) to[out=0,in=0] (edge_decoder.east);
    \path [line] (embedding.east) to[out=0,in=0]  (n_decoder_1.east);
    \path [line] (embedding.east) to[out=0,in=0] (n_decoder_more.east);
    \path [line] (edge_decoder) -- (inner_d_output);
    \path [line] (n_decoder_1) -- (node_d_output);
    \path [line] (n_decoder_more) -- (node_more_d_output);

    \path [line] (inner_d_output.west) to[out=180,in=0] (losses.east);
    \path [line] (node_d_output.west) to[out=180,in=0] (losses.east);
    \path [line] (node_more_d_output.west) to[out=180,in=0] (losses.east);

\end{tikzpicture}
    \caption{Graph Autoencoder approach.}
    \label{fig:GAE_workflow}
\end{figure}
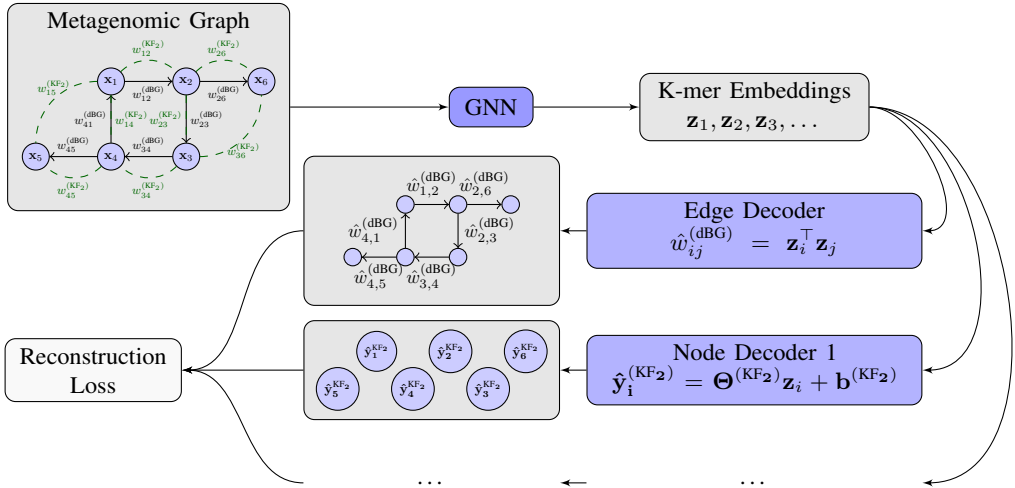

\paragraph{Edge Decoder}
The edge decoder's role is to capture contextual relationships from the original De Bruijn graph by reconstructing its transition probabilities. We employ an inner product decoder, defined as,
\[
    \hat{w}_{ij}^{(dBG)} = \mathbf{z}_i^T \mathbf{z}_j.
\]

\paragraph{Node Decoders}
The node decoder's role is to create embeddings reflecting structural similarities among k-mers by reconstructing the sub-k-mer frequency vectors for each node. Our model works with more than one sub\_k edge subtype, training a separate decoder for each. Each decoder consists of a linear layer, formulated as,
\[
    \mathbf{\hat{y}_i^{(\text{KF}_{\text{sub\_k}})}} 
    = 
    \mathbf{\Theta^{(\text{KF}_{\text{sub\_k}})}} \mathbf{z}_i 
    + 
    \mathbf{b^{(\text{KF}_{\text{sub\_k}})}},
\]
with separate parameters \(\mathbf{\Theta^{(\text{KF}_{\text{sub\_k}})}}\) and \(\mathbf{b^{(\text{KF}_{\text{sub\_k}})}}\) for each sub\_k.

\paragraph{Reconstruction Loss}
In our GAE model, the  L1 loss is used for the edge decoder, while the Mean Squared Error is applied to the node decoders. The total loss, denoting the set of all sub\_ks as \(K\) and the total number of nodes as \(N\), is calculated as follows,
\[
L_{\text{GAE}} = \sum_{(i,j) \in E} |w^{(dBG)}_{ij} - \hat{w}^{(dBG)}_{ij}| + \frac{1}{|K|} \sum_{\text{sub\_k}  \in K} \sum_{i \in \text{N}} (\mathbf{y^{(\text{KF}_{\text{sub\_k}} )}_i} - \mathbf{\hat{y}^{(\text{KF}_{\text{sub\_k}})}_i} )^2.
\]

\paragraph{Results} The results of applying the GAE to the Edit Distance Approximation and Closest String Retrieval tasks are presented in Tables \ref{tbl:gae} and \ref{tbl:gae_top}. Although the GAE demonstrates overall good performance, often surpassing Word2Vec and Node2Vec, Contrastive Learning consistently achieves superior results.

\begin{table}
\scriptsize
\centering
\caption{RMSE \(\downarrow\) for the Edit Distance Approximation Task fine-tuned with Single Linear Layer. The standard deviation is based on three runs.}
\label{tbl:gae}
\begin{tblr}{
  cells = {c},
  column{1} = {r},
  cell{1}{2} = {c=2}{},
  cell{1}{4} = {c=2}{},
  vline{2-3} = {1}{},
  vline{2,4} = {2-9}{},
  hline{1,10} = {-}{0.08em},
  hline{3} = {-}{},
}
    & RT988 dataset            &                          & Qiita dataset            &                          \\
$k$ & Our GAE                  & Our CL                   & Our GAE                  & Our CL                   \\
2   & $0.40 \pm 0.01$          & $0.40 \pm 0.01$          & $2.34 \pm 0.01$          & $\mathbf{2.14 \pm 0.02}$ \\
3   & $0.37 \pm 0.01$          & $0.37 \pm 0.01$          & $2.10 \pm 0.03$          & $\mathbf{2.09 \pm 0.03}$ \\
4   & $0.37 \pm 0.01$          & $0.37 \pm 0.01$          & $2.09\pm 0.01$           & $\mathbf{2.00 \pm 0.01}$ \\
5   & $0.36 \pm 0.01$          & $\mathbf{0.35 \pm 0.01}$ & $2.09 \pm 0.03$          & $\mathbf{2.00 \pm 0.01}$ \\
6   & $\mathbf{0.35 \pm 0.01}$ & $0.36 \pm 0.01$          & $2.05 \pm 0.05$          & $\mathbf{1.97 \pm 0.01}$ \\
7   & $0.36 \pm 0.01$          & $0.36 \pm 0.01$          & $\mathbf{1.96 \pm 0.02}$ & $1.99 \pm 0.01$          \\
8   & $0.36 \pm 0.01$          & $\mathbf{0.35 \pm 0.01}$ & $2.06 \pm 0.01$          & $\mathbf{1.96 \pm 0.01}$ 
\end{tblr}
\end{table}

\begin{table}[!h]
\centering
\caption{Mean Top retrieval performance. The standard deviation is based on three runs.}
\label{tbl:gae_top}
\begin{subtable}{\textwidth}
\caption{Top 1\% $\uparrow$}
\scriptsize
\centering
\setlength{\extrarowheight}{0pt}
\addtolength{\extrarowheight}{\aboverulesep}
\addtolength{\extrarowheight}{\belowrulesep}
\setlength{\aboverulesep}{0pt}
\setlength{\belowrulesep}{0pt}

\begin{tabular}{r|cc|cc} 
\toprule
    & \multicolumn{2}{c|}{Zero-Shot: Aggregated K-mer Embeddings} & \multicolumn{2}{c}{Fine-Tuned: \textit{NeuroSEED} with K-mer Embeddings}  \\ 
\cline{2-5}
$k$ & Our GAE                  & Our CL                            & Our GAE         & Our CL                                                   \\ 
\hline
2   & $47.4 \pm 0.6$          & $\mathbf{52.2 \pm 0.7}$           & $47.9 \pm 1.1$ & $48.3 \pm 0.7$                                           \\
3   & $50.1 \pm 0.3$          & $\mathbf{53.1 \pm 0.4}$           & $47.9 \pm 0.7$ & $48.2 \pm 0.3$                                           \\
4   & $49.4 \pm 0.5$          & $\mathbf{53.3 \pm 0.3}$           & $49.3 \pm 0.5$ & $47.7 \pm 0.8$                                           \\
5   & $50.0 \pm 0.2$          & $\mathbf{50.5 \pm 0.1}$           & $49.4 \pm 0.2$ & $47.8 \pm 0.3$                                           \\
6   & $\mathbf{50.2 \pm 0.7}$ & $50 \pm 0.7$                      & $48.5 \pm 1.4$ & $47 \pm 0.9$                                             \\
7   & $\mathbf{49.8 \pm 0.3}$ & $48.3 \pm 1.1$                    & $49.4 \pm 0.7$ & $48.9 \pm 0.6$                                           \\
8   & $45.0 \pm 0.1$          & $\mathbf{50.2 \pm 0.1}$           & $49.4 \pm 1.2$ & $48 \pm 0.3$                                             \\
\bottomrule
\end{tabular}
\end{subtable}
~
\begin{subtable}{\textwidth}
\caption{Top 10\% $\uparrow$}
\footnotesize
\scriptsize
\centering
\setlength{\extrarowheight}{0pt}
\addtolength{\extrarowheight}{\aboverulesep}
\addtolength{\extrarowheight}{\belowrulesep}
\setlength{\aboverulesep}{0pt}
\setlength{\belowrulesep}{0pt}
\begin{tabular}{r|cc|cc} 
\toprule
    & \multicolumn{2}{c|}{Zero-Shot: Aggregated K-mer Embeddings} & \multicolumn{2}{c}{\begin{tabular}[c]{@{}c@{}}Fine-Tuned: \textit{NeuroSEED} with K-mer Embeddings\end{tabular}}  \\ 
\cline{2-5}
$k$ & Our GAE                  & Our CL                            & Our GAE         & Our CL                                                                                                                                           \\ 
\hline
2   & $62.6 \pm 0.2$          & $68.0 \pm 1.1$                    & $75.4 \pm 0.3$ & $\mathbf{76.4 \pm 0.5}$                                                                                                                          \\
3   & $68.1 \pm 0.4$          & $70.8 \pm 0.6$                    & $75.4 \pm 0.8$ & $\mathbf{75.6 \pm 0.6}$                                                                                                                          \\
4   & $75.1 \pm 0.3$          & $\mathbf{78.1 \pm 0.1}$           & $75.3 \pm 0.3$ & $75.4 \pm 0.3$                                                                                                                                   \\
5   & $78.5 \pm 0.5$          & $\mathbf{79.5 \pm 0.2}$           & $75.4 \pm 119$ & $75.3 \pm 0.8$                                                                                                                                   \\
6   & $80.4 \pm 0.4$          & $\mathbf{81.3 \pm 0.1}$           & $75.2 \pm 0.6$ & $74.8 \pm 0.7$                                                                                                                                   \\
7   & $\mathbf{81.3 \pm 0.5}$ & $80.1 \pm 0.5$                    & $75.6 \pm 1.0$ & $77.8 \pm 0.5$                                                                                                                                   \\
8   & $70.9 \pm 0.2$          & $\mathbf{79.4 \pm 0.2}$           & $76.5 \pm 0.5$ & $75.4 \pm 0.4$                                                                                                                                   \\
\bottomrule
\end{tabular}
\end{subtable}
\vspace{-2em}
\end{table}

\end{document}